\documentclass[10pt,twocolumn,letterpaper]{article}

\usepackage{iccv}
\usepackage{times}
\usepackage{epsfig}
\usepackage{graphicx}
\usepackage{amsmath}
\usepackage{amssymb}
\usepackage{graphicx}
\usepackage{amsmath}
\usepackage{hhline}
\usepackage{pifont}
\newcommand{\cmark}{\ding{51}}%
\newcommand{\xmark}{\ding{55}}%
\usepackage{multirow}
\usepackage{authblk}
\usepackage{amsmath,amssymb}

\usepackage[pagebackref=true,breaklinks=true,letterpaper=true,colorlinks,bookmarks=false]{hyperref}

\iccvfinalcopy 

\def\iccvPaperID{****} 

\ificcvfinal\fi

\begin{document}

\title{UPI-Net: Semantic Contour Detection in Placental Ultrasound}
\author[1]{Huan Qi}
\author[2]{Sally Collins}
\author[1]{J. Alison Noble}
\affil[1]{Institute of Biomedical Engineering, Department of Engineering Science, University of Oxford}
\affil[2]{Nuffield Department of Women's and Reproductive Health, University of Oxford}

\maketitle
\ificcvfinal\thispagestyle{empty}\fi

\begin{abstract}
Semantic contour detection is a challenging problem that is often met in medical imaging, of which placental image analysis is a particular example. In this paper, we investigate utero-placental interface (UPI) detection in 2D placental ultrasound images by formulating it as a semantic contour detection problem. As opposed to natural images, placental ultrasound images contain specific anatomical structures thus have unique geometry. We argue it would be beneficial for UPI detectors to incorporate global context modelling in order to reduce unwanted false positive UPI predictions. Our approach, namely UPI-Net, aims to capture long-range dependencies in placenta geometry through lightweight global context modelling and effective multi-scale feature aggregation. We perform a subject-level 10-fold nested cross-validation on a placental ultrasound database (4,871 images with labelled UPI from 49 scans). Experimental results demonstrate that, without introducing considerable computational overhead, UPI-Net yields the highest performance in terms of standard contour detection metrics, compared to other competitive benchmarks.

\end{abstract}

\section{Introduction}
\label{intro}
Placenta accreta spectrum (PAS) disorders denote a variety of adverse pregnancy conditions that involve abnormally adherent or invasive placentas towards the underlying uterine wall. Without risk assessment, any attempt to remove the embedded organ may cause catastrophic maternal haemorrhage \cite{jauniaux2018placenta}. Reduction of maternal mortality and morbidity of PAS disorders relies on both recognition of women at risk and more importantly, on accurate prenatal diagnosis. However, recent population studies have shown unsatisfactory results: PAS disorders remain undiagnosed before delivery in one-third to two-thirds of cases \cite{jauniaux2018figo2}. Over the last 40 years, a 10-fold increase in the incidence of PAS disorders has been reported in most medium- and high-income countries with the rising of cesarean delivery rates \cite{jauniaux2018figo2}. 

\begin{figure}[!t]
	\centering
	{\includegraphics[width=1\linewidth]{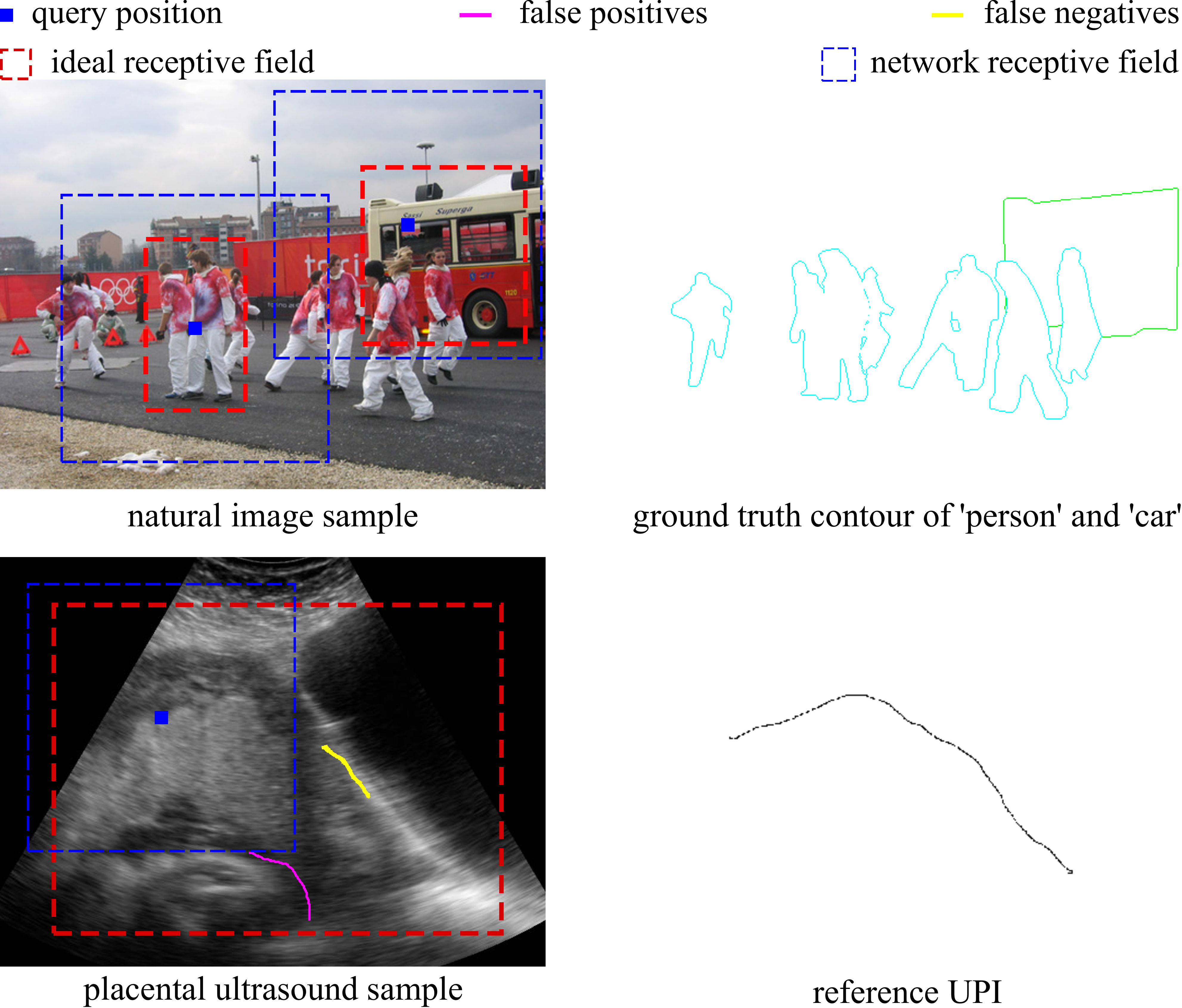}}
	\caption{Semantic contour detection in natural images (sample from SBD) and placental ultrasound images. Best viewed in color.} 
	\label{fig:1}
\end{figure}

\begin{figure*}[!t]
	\centering
	{\includegraphics[width=1\linewidth]{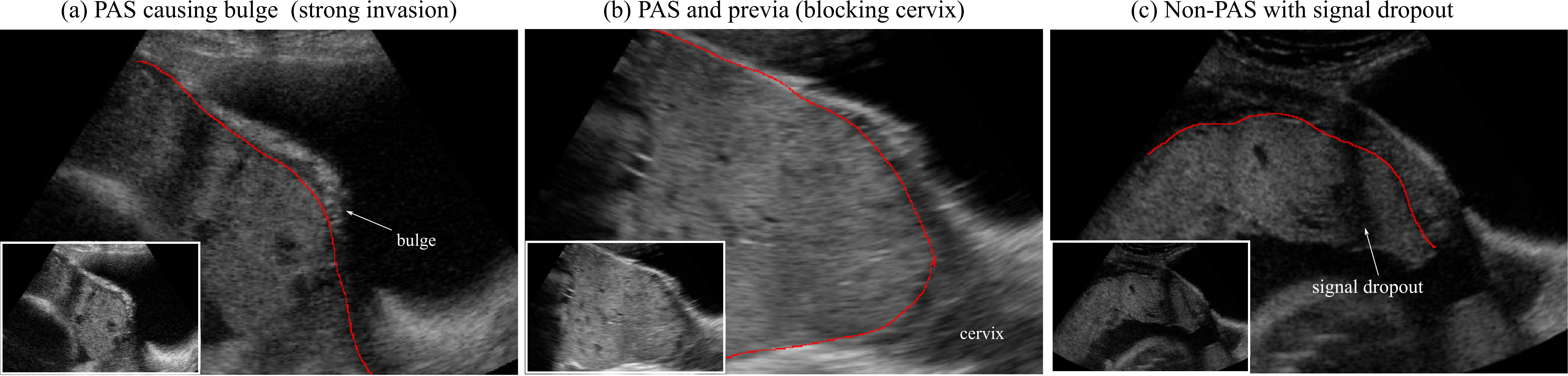}}
	\caption{Utero-placental interfaces (UPI) are annotated as red curves in placental ultrasound image samples. In PAS cases, myometrium tends to disappear due to placental invasion, causing weaker contrast around the UPI. Strong placental invasion would cause bulge-like UPI as shown in (a). In placenta previa case, UPI usually takes a `U-shape' over cervix, as shown in (b). A non-PAS case where UPI separates the placenta from the myometrium is shown in (c), somewhat contaminated by signal dropout.} 
	\label{fig:2}
\end{figure*}

Ultrasonography is widely used to assist diagnosis of PAS disorders prenatally. Recently, the International Federation of Gynecology and Obstetrics released consensus guidelines on PAS disorders in terms of prenatal diagnosis and screening \cite{jauniaux2018figo2}, among which identifying structural and vascular abnormalities near the utero-placental interface (UPI) is of key importance. UPI is the anatomical interface that separates the placenta from the uterus. In non-PAS cases, the UPI is observed as the placental boundary that touches the myometrium. However, in PAS cases, the degree of placental invasion can vary along the UPI, resulting in an irregular shape and length with low contrast. Manual localization remains challenging and time-consuming even for experienced sonographers, as shown in Fig.~\ref{fig:1} and Fig.~\ref{fig:2}. 

In order to recognize edge pixels of specific semantic categories, convolution neural networks are often designed to have large receptive fields by repeatedly stacking downsampling and (dilated) convolution layers \cite{he2015delving,he2016deep,yu2017dilated,yu2015multi}, which is reported to be computationally inefficient and difficult to optimize in general \cite{wang2018non,cao2019gcnet}. To address this issue, a self-attention mechanism, originally born in natural language processing studies \cite{vaswani2017attention}, can be introduced to explicitly model element-wise correlation \cite{zhang2018self, wang2018non} and has achieved success in video classification, object detection and segmentation \cite{hu2018relation,zhang2018context}. 

Fig.~\ref{fig:1} displays two sample images from the Semantic Boundaries Dataset (SBD) and our PAS database respectively. In natural images, objects of interest may appear with various scales at different locations within a scene. More often than not, the network receptive field is large enough to capture relevant semantics for semantic contour detection. On the contrary, placental ultrasound images contain specific anatomical structures thus have unique geometry. From a low-level perspective, there is a considerable amount of UPI-like edges (false positives, e.g. in Fig.~\ref{fig:1}). We need to suppress irrelevant edges that are not UPI (i.e. do not separate the placenta from the uterus) by modelling high-level semantics, which requires the network to also identify specific semantic entities related to placenta geometry \cite{li2019spatial,sabour2017dynamic}. Moreover, we observe false negatives in some low-contrast regions. We expect to alleviate these errors by incorporating long-range contextual cues \cite{wang2018non,cao2019gcnet}. To this end, we argue that it would be beneficial for UPI detectors to model global context of each spatial position in order to suppress false predictions thus improve detection performance. 

In this paper, we propose UPI-Net, a deep network designed for UPI detection in placental ultrasound images, as a critical step in an image-based PAS prenatal diagnosis pipeline. UPI-Net captures the long-range dependencies in placenta geometry using lightweight global context modelling units and effective multi-scale feature aggregation. The contributions are twofold. First, we propose a novel architecture to enforce contextual feature learning in earlier stages and enhance learning of UPI-related semantic entities / geometry in later stages. Second, we demonstrate the effectiveness of UPI-Net by comparing against several competitive benchmarks on a placental ultrasound database. Performances of UPI detectors are evaluated using standard edge/contour detection metrics \cite{arbelaez2011contour,hariharan2011semantic}. According to experiments, UPI-Net yields the best performance without introducing considerable computational overhead.

\begin{figure*}[!t]
	\centering
	{\includegraphics[width=1\linewidth]{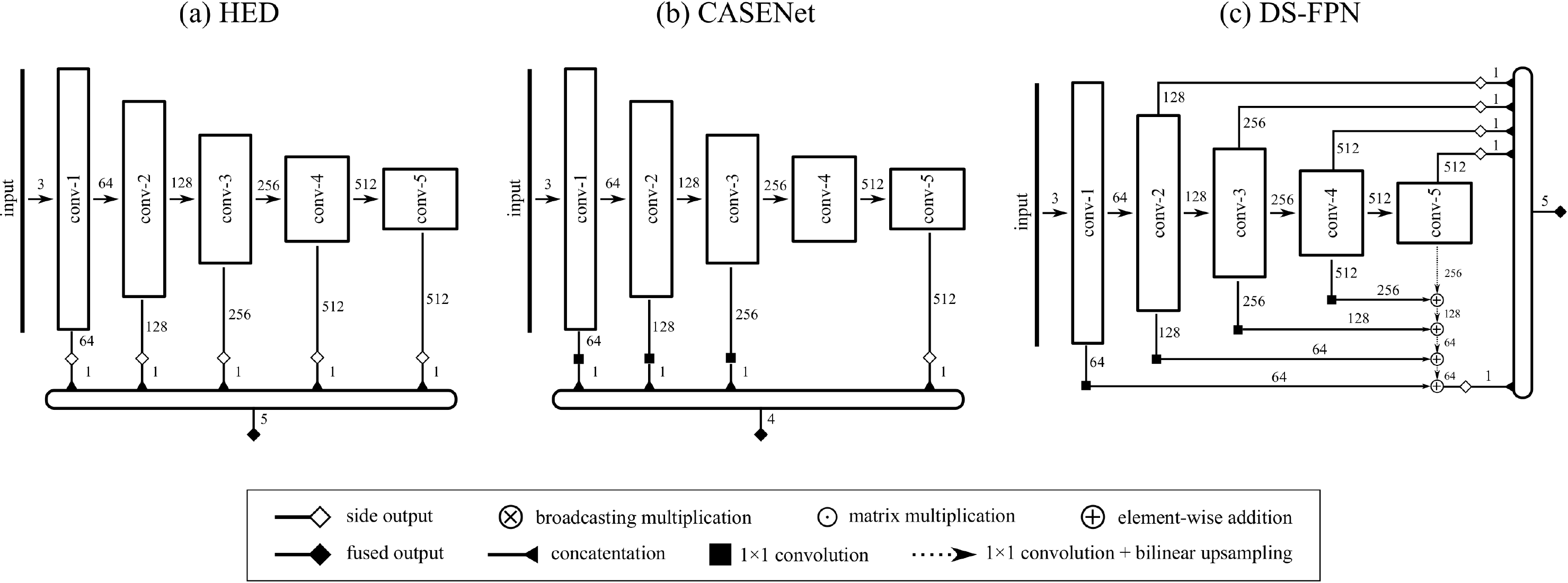}}
	\caption{Three multi-scale feature aggregation architectures: (a) HED \cite{xie2015holistically}; (b) CASENet \cite{yu2017casenet}; (c) DS-FPN \cite{qi2018automatic}.} 
	\label{fig:3}
\end{figure*} 
\section{Related Work}

\noindent{\bf Semantic contour detection.} Edge detection is one of the fundamental tasks in computer vision and has been extensively studied in the past. However, assigning semantics to edges is a relatively new task that has not received much attention in both natural image and medical image analysis \cite{merkow2016dense,hariharan2011semantic,aslam2015improved}. Early work uses class-specific edges for tracking \cite{shahrokni2005classifier,dollar2006supervised}, object detection and segmentation \cite{prasad2006learning}. Hariharan et al. presented the large-scale Semantic Boundaries Dataset (SBD) and proposed to use generic object detectors along with bottom-up contours for semantic contour detection \cite{hariharan2011semantic}. Bertasius et al. introduced a CNN-based two-stage process that first identified all edge candidates and then classified them using segmentation networks \cite{bertasius2015high,long2015fully,chen2017deeplab}. Yu et al. proposed CASENet to detect semantic edges in an end-to-end fashion. They optimized the holistically-nested edge detection network (HED) \cite{xie2015holistically} by removing deep supervisions on the early-stage side outputs and instead using them as shared features for the final fusion \cite{yu2017casenet}. The proposed UPI-Net adopts a nested architecture as CASENet does but extends it by adding global context modelling units that are well-suited for UPI prediction.

\noindent{\bf Global context modelling.} Attention-based global context modelling has been successfully applied in various visual recognition applications such as semantic segmentation \cite{zhang2018context}, panoptic segmentation \cite{li2019attention}, video classification \cite{wang2018non}, generative adversarial networks \cite{zhang2018self}, and representation learning \cite{cao2019gcnet,hu2018squeeze,hu2018squeeze,li2019spatial,park2018bam,woo2018cbam,gao2019res2net}. It is recently reported that the non-local pixel-wise attention can be simplified as a more memory-efficient query-independent attention without sacrificing performance \cite{wang2018non,cao2019gcnet}. Following this work, UPI-Net models the global context of placental ultrasound images via lightweight non-local heads and semantic enhancement heads without introducing a large amount of network parameters or computational overhead.

\section{Methods}
\subsection{Problem Formulation}
\noindent{\bf Training process.} Our training set is denoted as $D=\{(\mathcal{X}_n, \mathcal{Y}_n),n=1,\cdots,|D|\}$, where a sample $\mathcal{X}_n=\{x_n^p, p=1,\cdots,|\mathcal{X}_n|\}$ denotes a placental ultrasound image and $\mathcal{Y}_n=\{y_n^p, p=1,\cdots,|\mathcal{X}_n|\}$ denotes the corresponding reference UPI map for $\mathcal{X}_n$. $\mathcal{Y}_n$ takes the form of a binary mask with $y_n^p\in\{0,1\}$, i.e. pixels on the UPI take the value 1. For notation simplicity, we drop the subscript $n$ from now on. Our goal is to train a network with parameters $\mathbf{W}$ to predict the probability $Pr(y^p=1|\mathcal{X}; \mathbf{W})$ at each pixel position $p$ in $\mathcal{X}$. Following \cite{xie2015holistically,yu2017casenet}, we introduce a class-balancing weight $\omega$ to alleviate the extremely low foreground-background class ratio encountered during training. This is based on the idea of prior scaling \cite{lawrence1998neural}, with the purpose to equalize the expected model weight update for both classes. Specifically, we define the following cross-entropy loss function on the network output $\mathcal{O}$ given a training pair $(\mathcal{X}, \mathcal{Y})$:
\begin{align*}
L(\mathcal{O}; \mathbf{W}) &= -\frac{\omega}{|\mathcal{Y}_+|+|\mathcal{Y}_-|}\sum_{p\in|\mathcal{Y}_+|}\log Pr(y^p=1|\mathcal{X}; \mathbf{W})\\ -&\frac{1}{|\mathcal{Y}_+|+|\mathcal{Y}_-|}\sum_{p\in|\mathcal{Y}_-|}\log(1-Pr(y^p=1|\mathcal{X}; \mathbf{W}))
\end{align*}
We set $\omega=\frac{|\mathcal{Y}_-|}{|\mathcal{Y}_+|}$, where $|\mathcal{Y}_+|$ and $|\mathcal{Y}_-|$ denote the number of positives and negatives. The network output $\mathcal{O}$ at pixel position $p$ is activated by a sigmoid function to obtain $Pr(y^p=1|\mathcal{X}; \mathbf{W})$:
\begin{align*}
Pr(y^p=1|\mathcal{X}; \mathbf{W}) = \frac{1}{1+\exp(-\mathcal{O}_p)}
\end{align*}
UPI-Net has two outputs, a side output $\mathcal{O}_s$ and a fused output $\mathcal{O}_f$. The details will be discussed in Sec.~\ref{arch}. Each output corresponds to an individual prediction. The overall loss function is simply the sum of losses on individual outputs:
\begin{align*}
L_{all}(\mathbf{W})&= L(\mathcal{O}_s; \mathbf{W})+L(\mathcal{O}_f; \mathbf{W})
\end{align*}

\noindent{\bf Testing process.} During testing, we obtain two outputs from UPI-Net given an unseen placental ultrasound image $\mathcal{X}$. The final prediction is simply the sigmoid of the fused output, i.e. $\frac{1}{1+\exp(-\mathcal{O}_f)}$.

\begin{figure*}[!t]
	\centering
	{\includegraphics[width=0.95\linewidth]{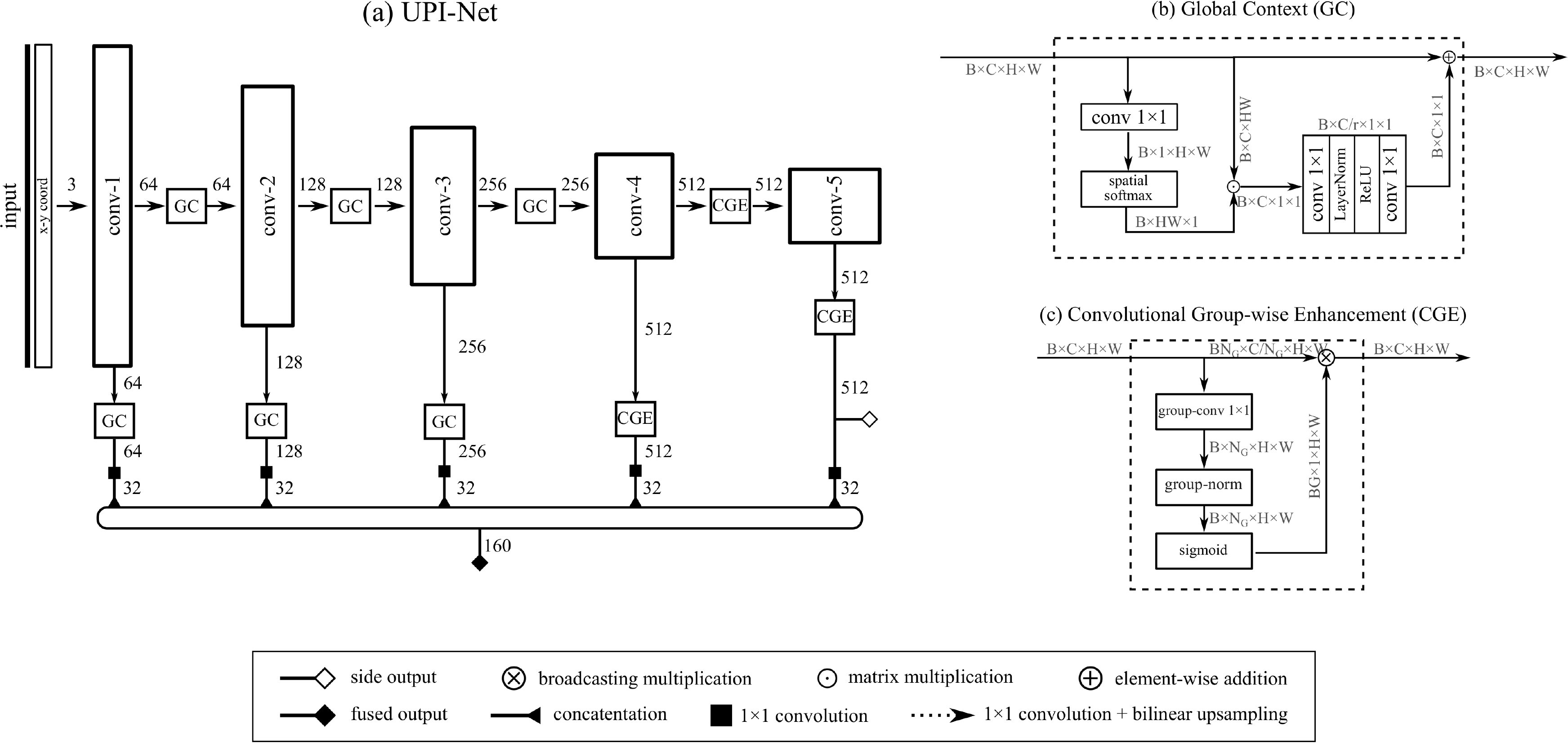}}
	\caption{(a) Proposed UPI detector layout, where an ImageNet-pretrained VGG-16 is the backbone; (b) A global context (GC) block; (c) A convolutional group-wise enhancement (CGE) block.} 
	\label{fig:4}
\end{figure*}

\subsection{Network Architecture}
\label{arch}
Rich hierarchical representations of deep neural networks lead to success in edge detection \cite{xie2015holistically,yu2017casenet}. This is particularly important for UPI detection, which requires effective aggregation of multi-scale features to localize edge pixels on the UPI and get rid of false positives using global context of placenta geometry. In this sub-section, we first present three alternative multi-scale feature aggregation architectures that have been successfully used in edge detection and key-point localization \cite{qi2018automatic,yu2017casenet,xie2015holistically,lin2017feature}. Then we discuss their suitability for UPI detection and propose UPI-Net in an effort to resolve some of these issues.

\noindent{\bf Multi-scale feature aggregation}. As shown in Fig.~\ref{fig:3}, we present three architectures that aggregate multi-scale features: HED \cite{xie2015holistically}, CASENet \cite{yu2017casenet}, and DS-FPN \cite{qi2018automatic}. They are all built upon the classic VGG-16 network to be structurally consistent. HED inherits the idea of deeply-supervised nets \cite{lee2015deeply} to produce five individual side outputs at different scales and another fused output via multi-scale feature concatenation. CASENet adopts a similar nested architecture but disables early-stage deep supervisions thus only produces one side output and one fused output. DS-FPN extends the idea of feature pyramid networks \cite{lin2017feature} by connecting multi-scale features via $1\times1$ convolutions and element-wise additions, producing five side outputs and one fused output.

UPI detection depends both on low-level features associated with edges, which are well preserved in the shallower stages of the network, and on high-level semantic entities associated with placenta geometry, which are learnt in the deeper stages of the networks. One common issue related to the three architectures above is the sub-optimal use of low-level features. Previous work tends to use them for feature augmentation without careful refinement. We believe it is beneficial for UPI detectors to incorporate global context modelling in features of different scales (esp. those in the shallower stages). Moreover, large receptive fields are only available in the deepest stages of the networks via stacked convolutional operations, which might not even be large enough to model important long-distance dependencies in placental ultrasound images, as discussed in Sec.~\ref{intro}. 
\begin{figure*}[!t]
	\centering
	{\includegraphics[width=1\linewidth]{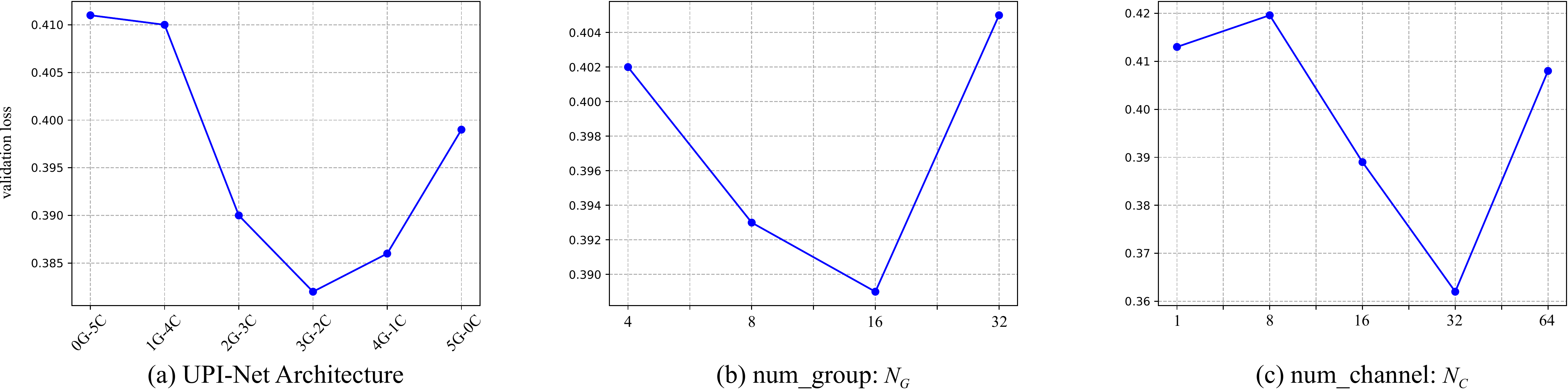}}
	\caption{Hyper-parameter searching for UPI-Net, where an iterative strategy is applied for better efficiency.} 
	\label{fig:5}
\end{figure*}
\noindent{\bf GC blocks}. Our proposed UPI-Net (Fig.~\ref{fig:4}) aims to address these potential issues by adding two types of feature refinement blocks in a nested deep architecture: (i) global context (GC) blocks \cite{cao2019gcnet}; (ii) convolutional group-wise enhancement (CGE) blocks. A GC block modulates low-level features via simplified non-local operations and channel recalibration operations. As shown in Fig.~\ref{fig:4}(b), it first performs global attention pooling on the input feature maps via a $1\times1$ convolution and a spatial softmax layer. The output is then multiplied with the original input to obtain a channel attention weight. After a channel recalibration transform (via $1\times1$ convolutions, $r=16$ \cite{hu2018squeeze}), the calibrated weight is aggregated back to the original input via a broadcasting addition. As reported in \cite{cao2019gcnet}, a GC block is a lightweight alternative to the non-local block \cite{wang2018non} in modelling global context of the input feature map. In UPI-Net, we attach GC blocks to {\it conv-1}, {\it conv-2} and {\it conv-3} to refine features from the earlier stages of the network.

\noindent{\bf CGE blocks}. Inspired by \cite{li2019spatial}, we introduce a convolutional group-wise enhancement (CGE) block to promote learning of high-level semantic entities related to UPI detection via group-wise operations. As shown in Fig.~\ref{fig:4}(c), a CGE block contains a group convolution layer (num$\_$group$=N_G$), a group-norm layer \cite{wu2018group}, and a sigmoid function. The group convolution layer essentially splits the input feature maps $B\times C\times H\times W$ into $N_G$ groups along the channel dimension. After convolution, each group contains a feature map of size $B\times1\times H\times W$. The subsequent group-norm layer normalizes each map over the space respectively. The learnable scale and shift parameters in group-norm layers are initialized to ones and zeros following \cite{wu2018group}. The sigmoid function serves as a gating mechanism to produce a group of {\it importance maps}, which are used to scale the original inputs via the broadcasting multiplication. We expect the group-wise operations in CGE to produce unique semantic entities across groups. The group-norm layer and sigmoid function can help enhance UPI-related semantics by suppressing irrelevant noise. Our proposed CGE block is a modified version of the spatial group-wise enhance (SGE) block in \cite{li2019spatial}. We replace the global attention pooling with a simple $1\times1$ group convolution as we believe learnable weights are more expressive than weights from global average pooling in capturing high-level semantics. Our experiments on the validation set empirically support this design choice. CGE blocks are attached to {\it conv-4} and {\it conv-5} respectively, where high-level semantics are learnt. 

\noindent{\bf UPI-Net}. All refined features are linearly transformed (num$\_$channel$=N_C$) and aggregated via channel-wise concatenation to produce the fused output. Additionally, we produce a side output using {\it conv-5} features, which encodes strong high-level semantics. As displayed in Fig.~\ref{fig:4}, channel mismatches are resolved by $1\times1$ convolution and resolution mismatches by bilinear upsampling. Furthermore, we add a Coord-Conv layer \cite{liu2018intriguing} in the beginning of the UPI-Net, which simply requires concatenation of two layers of coordinates in $(x,y)$ Cartesian space respectively. Coordinates are re-scaled to fall in the range of $[-1, 1]$. We expect that the Coord-Conv layer would enable the implicit learning of placenta geometry, which by the way does not add computational cost to the network. Experimental results on hyper-parameter tuning are presented in Sec.~\ref{exp:hyper}.

\begin{figure*}[!t]
	\centering
	{\includegraphics[width=0.9\linewidth]{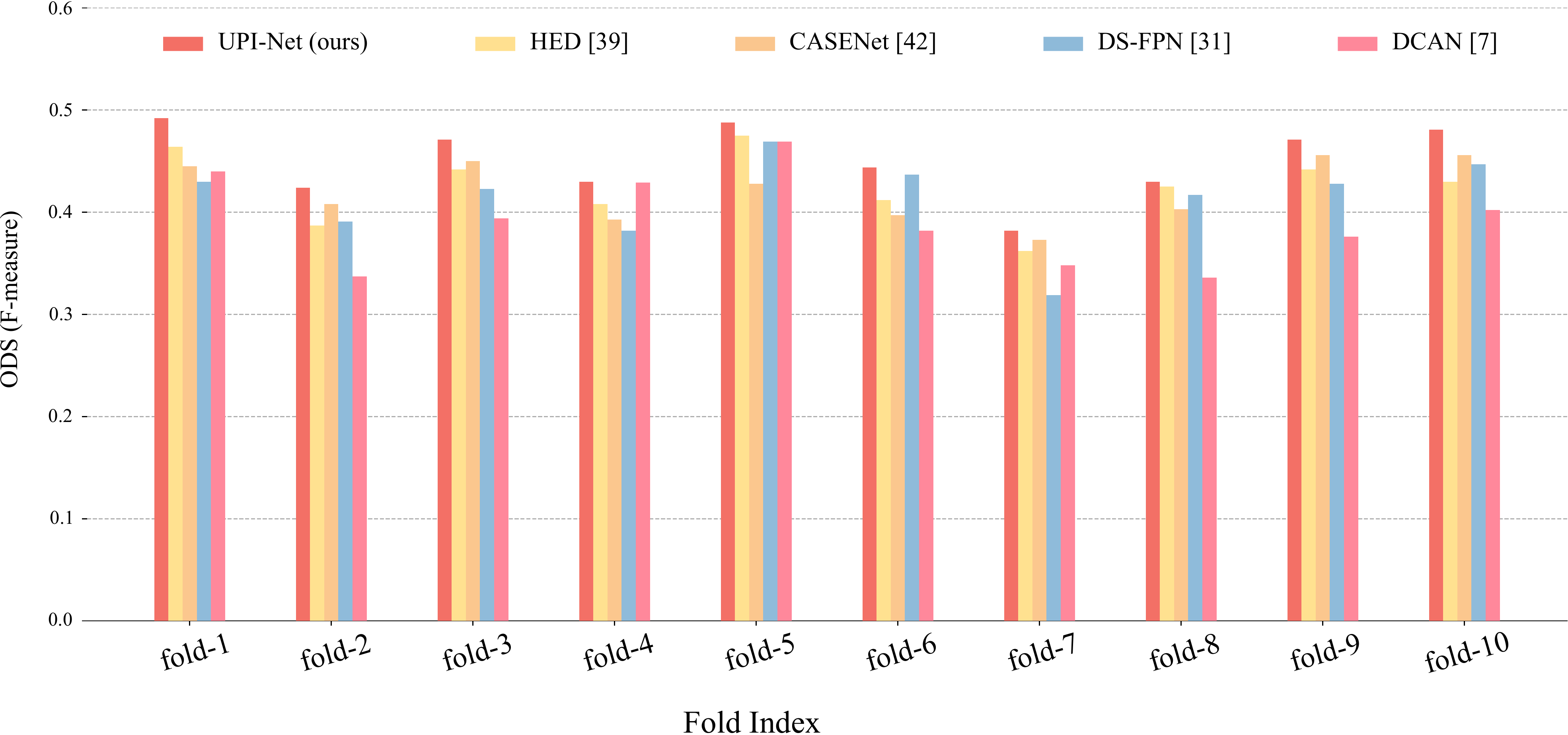}}
	\caption{Fold-wise performance comparison among UPI detectors.} 
	\label{fig:6}
\end{figure*}

\begin{table*}[]
	\centering
	\caption{The performance of different UPI detects on the test sets in a nested 10-fold cross validation. All results are in the format of {\it median [first, third quartile]}. $\downarrow_0$ indicates a lower value is more appreciated, with 0 being the best in theory. $\uparrow_1$ indicates a higher value is more appreciated, with 1 being the best in theory. ODS is the primary metric.}
	\label{tab:1}
	\begin{tabular}{l|c|c|c|c}
		\hline
		Model                     & Params (M)            & FLOPs (G)             & ODS  $\uparrow_1$                     & OIS $\uparrow_1$                     \\\hhline{=====}
		HED \cite{xie2015holistically}               & 14.7                  & 52.3                  & 0.427 {[}0.409, 0.442{]} & 0.469 {[}0.445, 0.487{]} \\
		CASENet \cite{yu2017casenet}              & 14.7                  & 52.3                  & 0.418 {[}0.399, 0.449{]} & 0.460 {[}0.442, 0.488{]} \\
		DS-FPN \cite{qi2018automatic}               & 15.1                  & 56.2                  & 0.426 {[}0.398, 0.435{]} & 0.465 {[}0.442, 0.480{]} \\
		DCAN \cite{chen2016dcan}                     & 8.6                   & 12.1                  & 0.388 {[}0.355, 0.422{]} & 0.439 {[}0.407, 0.473{]} \\\hline
		UPI-Net (ours)          &   14.7                  &       53.5                & {\bf 0.458 {[}0.430, 0.479{]}} & {\bf 0.493 {[}0.474, 0.518{]}} \\\hline
	\end{tabular}
\end{table*}

\section{Experiments}
\subsection{Dataset}
We had available 49 three-dimensional placental ultrasound scans from 49 subjects (31 PAS and 18 non-PAS) as part of a large obstetrics research project \cite{collins2012influence}. Written consents for obtaining the data was approved by the appropriate local research ethics committee. Static transabdominal 3D ultrasound volumes of the placental bed were obtained according to the predefined protocol with subjects in semi-recumbent position and a full bladder using a 3D curved array abdominal transducer. Each 3D volume was sliced along the sagittal plane into 2D images and annotated by X (a computer scientist) under the guidance of Y (an obstetric specialist). Unlike semantic contours in natural images, a UPI is characterized by low contrast, variable shape and signal attenuation. For manual annotation, human experts tend to rely on global context to first identify the UPI neighbourhood and then delineate it according to local cues. Due to the muscular nature of the uterus, the UPI would normally appear to be a smooth curve in placental ultrasound images, except when placental invasion penetrates muscle layers in the case of PAS disorders. The database contains 4,871 2D images in total, from 28 to 136 slices per volume with a median of 104 slices per volume.

\subsection{Evaluation protocol}
For a medical image analysis application with a relatively small dataset, a {\it non-nested k-fold cross-validation} is often used to compensate for the lack of test data (e.g. \cite{wang2019deep,gibson2018automatic,cciccek20163d,novikov2018fully}). However, this can lead to over-fitting in model selection and subsequent selection bias in performance evaluation \cite{cawley2010over}, causing overly-optimistic performance score for all the evaluated models. To avoid this problem, we carry out model selection and performance evaluation under a {\it nested} 10-fold cross-validation. Specifically, we run a 10-fold subject-level split on the database. In each fold, test data consisting of 2D image slices from 4 - 5 volumes are {\it held out}, while images from the remaining 44-45 volumes are further split into train/validation sets. In the inner loop (i.e. within each fold), we fit models to the training set and tune hyper-paramters over the splitted validation set. In the outer loop (i.e. across folds), generalization error is estimated on the held-out test set. We report evaluation scores on the test set splits to avoid potential {\it information leak}.

\subsection{Evaluation metrics}
Intuitively, UPI detection can be evaluated with standard edge detection metrics. We report two measures widely used in this field \cite{arbelaez2011contour}, namely the best F-measure on the dataset for a fixed prediction threshold (ODS), and the aggregate F-measure on the dataset for the best threshold in each image (OIS). Following \cite{xie2015holistically,yu2017casenet,arbelaez2011contour}, we choose the ODS F-measure as the primary metric since it balances the use of precision and recall at a fixed threshold. 

\subsection{Hyperparameter tuning}\label{exp:hyper}
\noindent{\bf GC / CGE configuration.} In UPI-Net, we attach GC blocks to the first three convolution units (i.e. {\it conv-1}, {\it conv-2} and {\it conv-3}) and CGE blocks to the last two. This configuration is chosen in the hyper-parameter tuning. Intuitively, GC blocks enforce non-local dependency across low-level features while CGE blocks promote learning of high-level semantics. An optimal configuration that balances low-level and high-level representation learning is desired. To this end, we vary the number of GC and CGE blocks to obtain different network variants, using {\it mG-nC} to represent first $m$ convolution units equipped with GC blocks and last $n$ convolution units with CGE blocks. For example, the proposed UPI-Net is denoted as {\it 3G-2C}. Fig.~\ref{fig:5}(a) displays the validation losses for different GC / CGE configurations, where {\it 3G-2C} is selected. 

\begin{figure*}[!t]
	\centering
	{\includegraphics[width=0.95\linewidth]{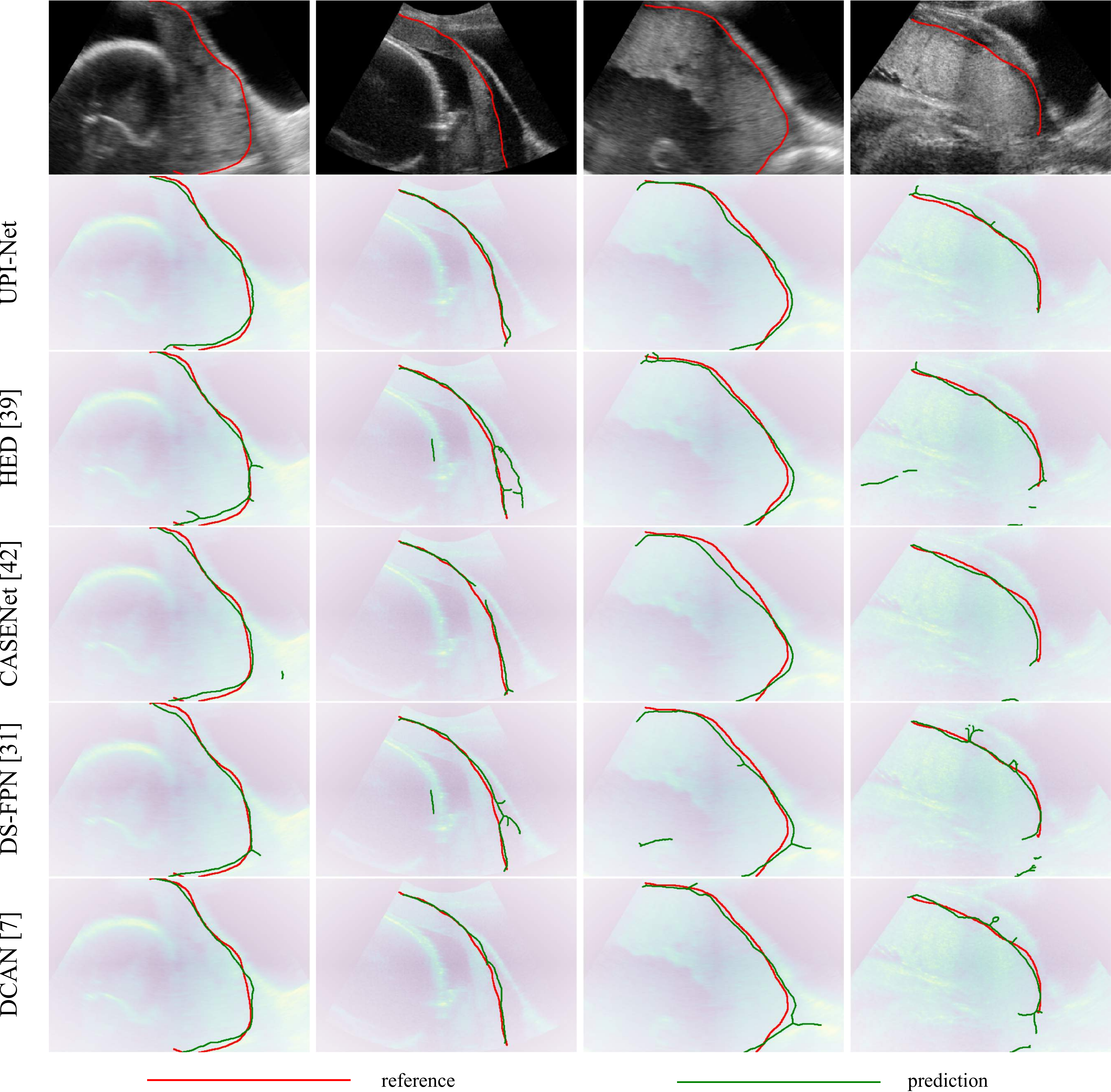}}
	\caption{Predictions from the proposed UPI-Net model and other benchmarks. UPI-Net suppresses a number of UPI-like false positives compared with other methods.} 
	\label{fig:7}
\end{figure*}

\noindent{\bf Group and aggregated feature channel number.} There are two more hyper-parameters introduced in Sec.~\ref{arch}, namely the number of groups ($N_G$) in CGE's group convolution layer and the number of channels ($N_C$) in the last feature aggregation layer. Similar to tuning the GC / CGE configuration, we vary $N_G$ and $N_C$ and test on the validation sets. Results are displayed in Fig.~\ref{fig:5}(b)-(c). Note that for simplicity, we fix the GC / CGE configuration as {\it 3G-2C} when searching for the optimial $N_G$ and then fix $N_G$ at the optimal value when searching for the optimal $N_C$. Such an iterative strategy efficiently reduces the hyper-parameter searching space. As a result, $N_G=16$ and $N_C=32$ are selected. It is noted that setting both hyper-parameters at larger values (e.g. $N_G=32$ and $N_C=64$)
does not necessarily reach a better performance for UPI detection.
\subsection{Implementation details}
Following the implementation details from the original papers, we used parameters from an ImageNet-pretrained VGG16 to initialize corresponding layers in HED, CASENet, DS-FPN and the proposed UPI-Net. Additionally, we implemented a DCAN model following the original design choice in \cite{chen2016dcan} without pretraining. The rest convolutional layers in UPI-Net were initialized by sampling from a zero-mean Gaussian distribution, following the method in \cite{he2015delving}. During training, we randomly cropped a patch of $320\times480$ px from the input images. For testing, we take the central crop of the same size. All inputs were normalized to have zero mean and unit variance. We used a mini-batch size of 8 to reduce memory footprint. With Adam optimizer, the initial learning rate was set to 0.0003. A weight-decay of 0.0002 was used. This hyper-paramater configuration was shared by all baseline models and UPI-Net variants.  All the models were implemented with PyTorch and trained for 40 epochs with early stopping on an NVIDIA DGX-1 with P100 GPUs.

\begin{figure}[!t]
	\centering
	{\includegraphics[width=\linewidth]{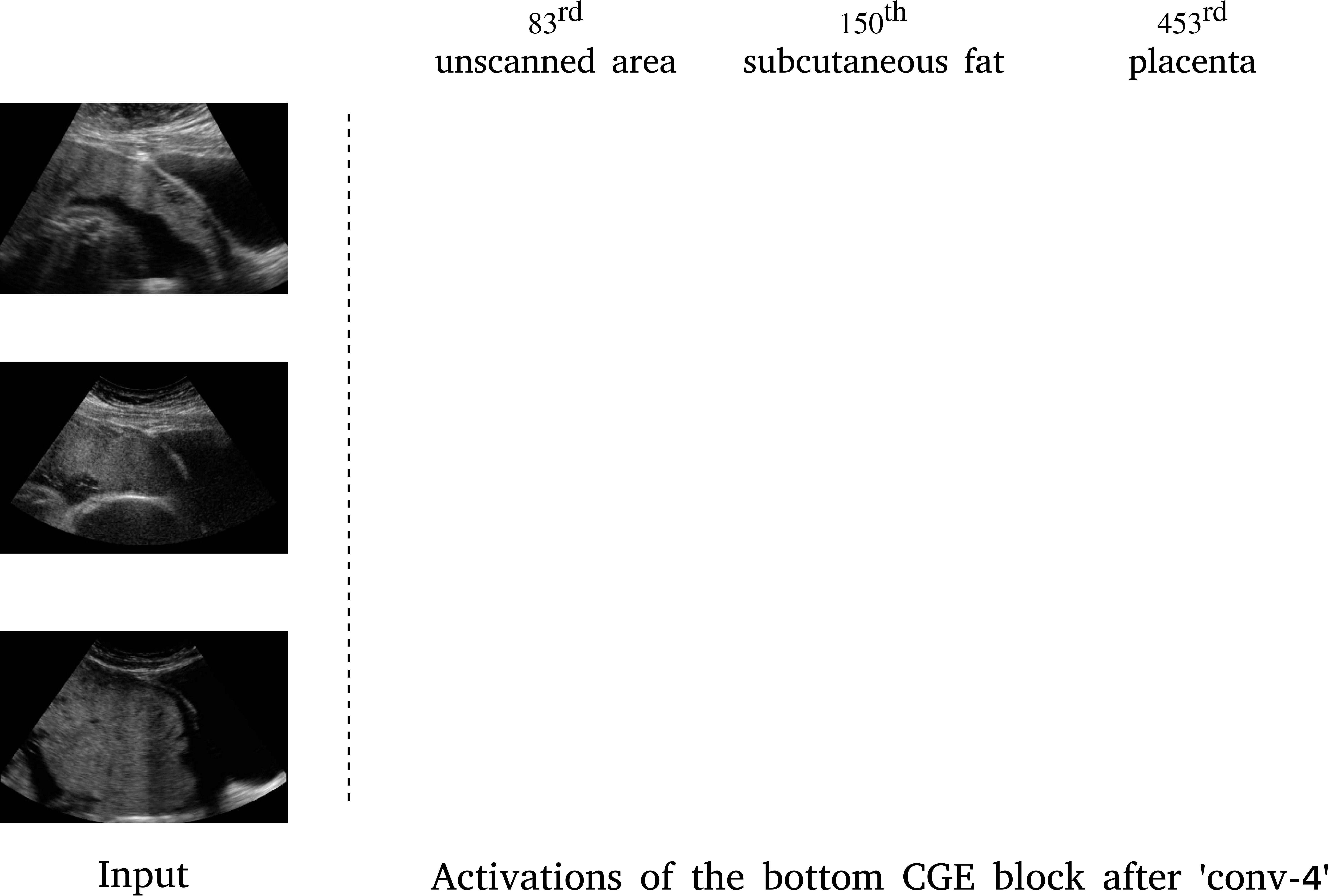}}
	\caption{Semantic entities learnt by UPI-Net.} 
	\label{fig:8}
\end{figure}

\subsection{Results}
Fold-wise performance comparison among UPI detectors are illustrated in Fig.~\ref{fig:6}. As shown in Table~\ref{tab:1}, the median, first and third quartile of the 10-fold test results are also presented. The proposed UPI-Net outperforms four competitive benchmarks in terms of ODS and OIS, without introducing a considerable amount of computational overhead in terms of model sizes and floating point operations. Test samples are displayed in Fig.~\ref{fig:7}. It is shown that predictions from UPI-Net are enhanced from a global perspective by suppressing unwanted UPI-like false positives and maintaining spatial smoothness of the curve (fewer false negatives).

\noindent{\bf Ablation study.} We further test how the Coord-Conv layer and the additional side-output supervision influence the performance of UPI detection. According to Table~\ref{tab:2}, UPI-Net benefits from both of them. Importantly, it costs no additional computational resources to add the Coord-Conv layer to the network. Although not used during testing, the side-output from {\it conv-5} modulates the training process to achieve better UPI detection.

\begin{table}[]
	\centering
	\caption{Ablation studies on Coord-Conv layers and the side-output supervision using {\it conv-5} features.}
	\label{tab:2}
	\begin{tabular}{l|c|c|c}
		\hline
		{\small Model}         & {\small Coord-Conv} & {\small Side-output} & {\small ODS$\uparrow_1$}                                         \\\hhline{====}
		Baseline-1     & \xmark          & \cmark           & {\scriptsize 0.438 {[}0.416, 0.463{]}}  \\
		Baseline-2     & \cmark          & \xmark           & {\scriptsize 0.444 {[}0.422, 0.454{]}}  \\\hline
		UPI-Net  & \cmark          & \cmark           & {\scriptsize 0.458 {[}0.430, 0.479{]}} \\\hline
	\end{tabular}
\end{table}

\noindent{\bf Learning semantic entities.} It is expected that introducing CGE modules would enable the network to learn high-level semantic entities related to placental geometry more effectively thus contributes to UPI detection. As displayed in Fig.~\ref{fig:8}, activation maps from the bottom CGE block after {\it conv-4} reveal some of the semantic entities learnt by UPI-Net. Particularly, the 453$^{rd}$ kernel appears to capture the placenta itself. Note that no supervision signal associated with the placenta location is available during training. This can be useful in clinical settings to assist operators in better interpreting the scene by visualizing regions of interest. 

\section{Conclusion}
We have presented a novel architecture for semantic contour detection for placental imaging. It can produce more plausible UPI predictions in terms of spatial continuity and detection performance via lightweight global contextual modelling, compared to competitive benchmarks. In addition to use in prenatal PAS assessment, we believe the proposed approaches could be adapted for other clinical scenarios that involves edge/contour detection in breast, liver, heart and brain imaging.

{\small
\bibliographystyle{ieee}
\bibliography{refs}
}

\end{document}